\documentclass[10pt,twocolumn,letterpaper]{article}
\usepackage[pagenumbers]{cvpr}             
\usepackage{xurl} 
\usepackage{mathptmx}
\usepackage{graphicx}
\usepackage{amsmath}
\usepackage{amssymb}
\usepackage{color}
\usepackage{bm}
\usepackage{multirow}
\usepackage{diagbox}
\usepackage{array}
\usepackage{booktabs}
\usepackage{nth}
\usepackage{nicefrac}
\usepackage{relsize}
\usepackage{xspace} 
\usepackage{bm}
\usepackage{enumerate}
\usepackage{enumitem}
\usepackage{array}
\usepackage{mathtools}

\usepackage{cuted}
\usepackage{capt-of}
\newcolumntype{P}[1]{>{\centering\arraybackslash}p{#1}}
\newcommand{\mytilde}{\raise.17ex\hbox{$\scriptstyle\mathtt{\sim}$}}
\usepackage[hang,flushmargin]{footmisc}
\newcommand{\bodymap}{BodyMap} 
\renewcommand{\thefootnote}{\fnsymbol{footnote}}

\newcommand\blfootnote[1]{%
  \begingroup
  \renewcommand\thefootnote{}\footnote{#1}%
  \addtocounter{footnote}{-1}%
  \endgroup
}

\newcommand{\beforefigcaption}{\vspace{-5.5mm}}
\newcommand{\afterfigcaption}{\vspace{-3mm}}
\newcommand{\beforetab}{\vspace{-3mm}}
\newcommand{\aftertab}{\vspace{-3mm}}

\usepackage[pagebackref,breaklinks=true,colorlinks,bookmarks=false]{hyperref}

\usepackage[capitalize]{cleveref}
\crefname{section}{Sec.}{Secs.}
\Crefname{section}{Section}{Sections}
\Crefname{table}{Table}{Tables}
\crefname{table}{Tab.}{Tabs.}

\def\cvprPaperID{3301} 
\def\confName{CVPR}
\def\confYear{2022}

\begin{document}
\title{BodyMap: Learning Full-Body Dense Correspondence Map}

\author{Anastasia Ianina\(^{1*}\), Nikolaos Sarafianos$^3$, Yuanlu Xu$^{3}$, Ignacio Rocco$^2$, Tony Tung$^3$\\
$^1$Moscow Institute of Physics and Technology, $^2$Meta AI, $^3$Meta Reality Labs Research, Sausalito\\
$^1${\tt\small yanina@phystech.edu}, $^{2,3}${\tt\small \{nsarafianos, yuanluxu, irocco, tonytung\}@fb.com}\\
\href{https://nsarafianos.github.io/bodymap}{\color{blue}https://nsarafianos.github.io/bodymap}
}

\maketitle

\begin{abstract}
\noindent Dense correspondence between humans carries powerful semantic information that can be utilized to solve fundamental problems for full-body understanding such as in-the-wild surface matching, tracking and reconstruction.
In this paper we present BodyMap, a new framework for obtaining high-definition full-body and continuous dense correspondence between in-the-wild images of clothed humans
and the surface of a 3D template model. The correspondences cover fine details such as hands and hair, while capturing regions far from the body surface, such as loose clothing.
Prior methods for estimating such dense surface correspondence
i) cut a 3D body into parts which are unwrapped to a 2D UV space, producing discontinuities along part seams, or
ii) use a single surface for representing the whole body, but none handled body details.
Here, we introduce a novel network architecture with Vision Transformers that learn fine-level features on a continuous body surface.
BodyMap outperforms prior work on various metrics and datasets, including DensePose-COCO by a large margin.
Furthermore, we show various applications ranging from multi-layer dense cloth correspondence, neural rendering with novel-view synthesis and appearance swapping.
\end{abstract}

\vspace{-2mm}
\section{Introduction}\label{sec:intro}

\begin{figure}[ptb]
    \vspace{-4mm}
    \centering
    \includegraphics[width=\linewidth]{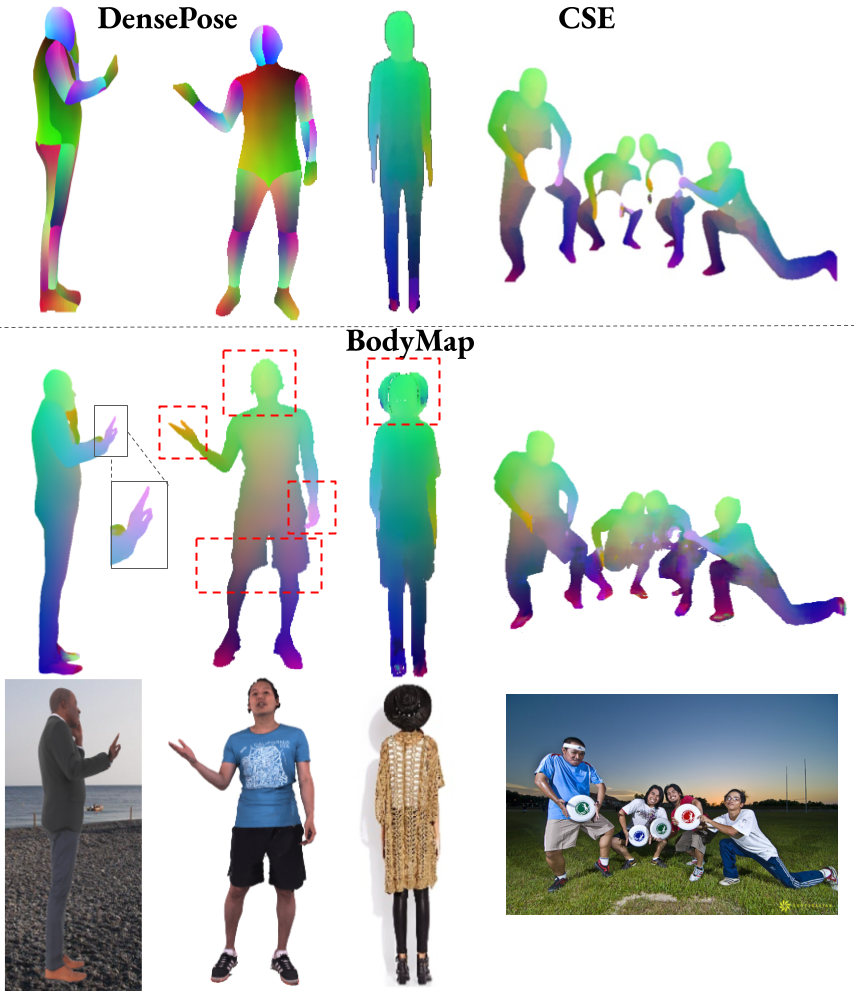}
    \beforefigcaption
    \caption{We introduce \textbf{BodyMap}~---~a method that establishes accurate dense correspondences between a 2D image and the surface of a 3D clothed human with high precision. Our approach handles loose clothes, different hairstyles and various accessories, like hats and bags, providing crisp silhouettes, and works well in multi-person cases with occlusions.
    }
    \afterfigcaption
    \label{fig:teaser}
\end{figure}
\blfootnote{*This work was conducted during an internship at Meta RL Research.}
\noindent Several fundamental problems related to human understanding in images can be addressed by labeling every pixel covering the human body with semantic information. This enables numerous applications including video analysis, image editing, texture generation and style transfer.
From a single RGB image of a human, the literature has proposed methods to extract sparse information such as 2D body keypoints (e.g., face, hands, body joints), or 2D segmentation masks (e.g., for full body, clothes, hair or skin), and also 3D body pose and shape parameters defined by a template body model~\cite{bogo2016keep, feng2021collaborative, pavlakos2019expressive, rong2020frankmocap, xu2019denserac, ExPose2020}, while work on dense surface correspondence has further enabled pixel-level understanding by establishing unique correspondences between 2D pixels covering the visible regions of the human body and 3D points on the surface of a body template.

In the seminal work DensePose~\cite{neverova2018dense}, correspondences are estimated between image pixels belonging to the human body and points in disjoint parts of a human body template located using UV coordinates, similar to local texture mapping.
The method is trained on the large in-the-wild dataset DensePose-COCO and is robust to human pose variability, image resolution, diversity in clothing, and occlusions. However, it has some inherent limitations that impact methods that rely on it (e.g., for clothed-human applications)~\cite{alldieck2019tex2shape, lazova2019360, yoon2021neural}. First, the discretization generated by dividing the body into disjoint parts produces clearly visible seams and discontinuities between them that are not optimal for learning models.
Second, the DensePose estimates suffer from inaccuracy as reported in prior work~\cite{neverova2018dense, neverova2019correlated, sarkarneural}, mainly due to the difficulty in acquiring ground-truth annotations for the task~\cite{alp2018densepose,lin2014microsoft}.
Follow-up methods have tackled some of its shortcomings and a few recent works addressed the discontinuity of UV maps~\cite{bhatnagar2020loopreg, neverova2020continuous, zeng20203d}.
HumanGPS~\cite{tan2021humangps} proposes to predict per-pixel embeddings using geodesic distances between corresponding points on the surface of a 3D human scan and does not produce an explicit mapping.
None of the proposed approaches has established high-definition correspondences for areas with finer details such as hair and hands (with fingers), with generalization to clothed humans, especially with loose clothing.

In this work we introduce a novel technique to establish high-definition full-body and continuous dense correspondence between images of \emph{clothed} humans and the human body surface.
Our method, which we term as \bodymap, takes as input an RGB image of a human and outputs accurate per-pixel continuous correspondence estimates for each foreground pixel (\ie including the full body, with clothes and hair).
We designed a transformer-based architecture that learns appearance-based and Continuous-Surface-Embeddings-based representations to infer accurate dense surface correspondence for the depicted human.
Our variant of Vision Transformer~\cite{dosovitskiy2020image} as a computational block of the encoder brings its advantageous properties for dense prediction tasks. The vector dimension is kept constant throughout all processing stages as well as global receptive field for every stage. With these properties, our network is well designed for dense correspondence prediction.

Furthermore, we capitalize on the power of synthetic data. Since no real-world dataset provides ground-truth annotation at the quality we aim for (fingers, clothes, hair), we created a synthetic dataset of animated 3D clothed human scans. In that way, we obtained ground-truth dense correspondence for a large variety of humans with diverse clothing, in different poses and from different viewpoints. A differentiating factor of our framework is that it is not tied to a human body with topology constraints, and can handle layered representations such as humans with separate cloth geometries. 
To summarize, our key contributions are:

\begin{itemize}[leftmargin=*]
\item \bodymap{} is the first method to establish dense continuous correspondence for every foreground pixel of clothed humans, whether that is fingers, hair, or clothes that are displaced from the human body with high-precision~---~something that all prior works fail to achieve. 
\item A novel transformer-based architecture designed specifically for this task that when trained in a multi-task learning manner with per-pixel classification losses for each channel significantly outperforms prior works across several datasets and tasks.
\item We achieve state-of-the-art results on DensePose COCO by a large margin. We show our approach can be applied to real-world applications such as novel view synthesis. Our method can be extended to learn layered representations with clothed humans and predict per-geometry surface correspondences.
\end{itemize}

\begin{figure*}[t]
    \vspace{-2mm}
    \centering
    \includegraphics[width=\linewidth]{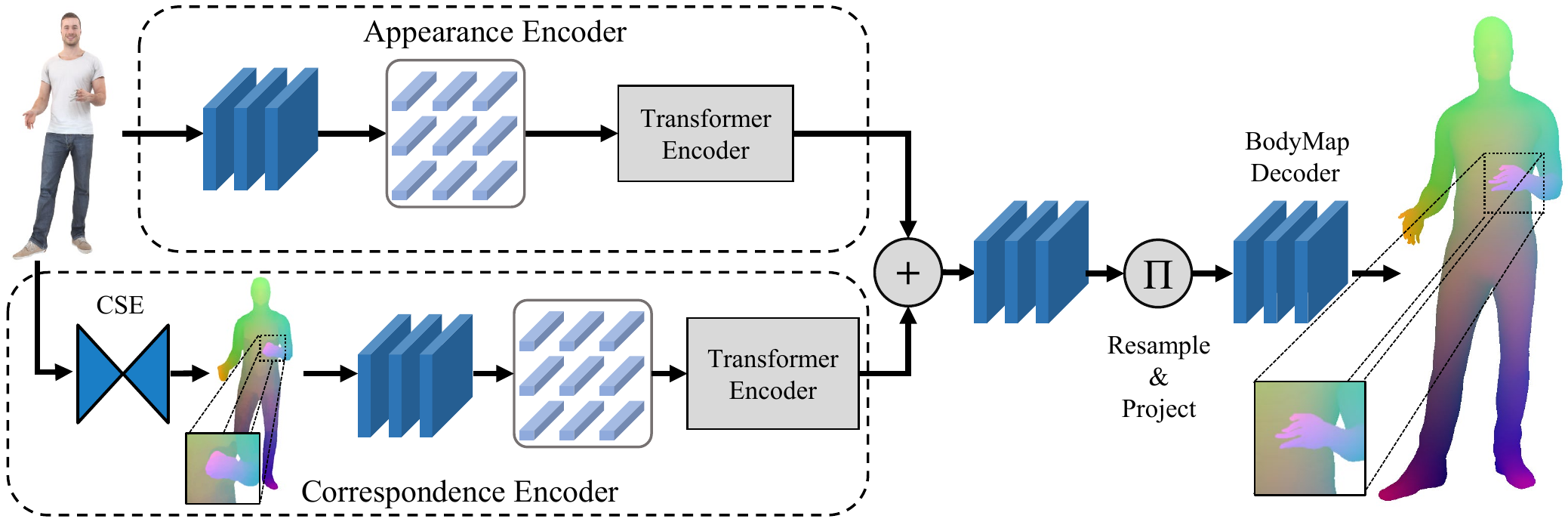}
    \beforefigcaption
    \caption{\textbf{BodyMap architecture}. Given an RGB image we first obtain its CSE~\cite{neverova2020continuous} estimates and feed both to their corresponding encoders. We utilize vision transformers specifically designed for this task to learn to extract accurate high-dimensional representations that are then fed to the BodyMap decoder that predicts per-pixel dense correspondences.}
    \afterfigcaption
    \vspace{-0.2cm}
    \label{fig:hd-cse}
\end{figure*}

\section{Related Work}\label{sec:related}
\noindent \textbf{Dense Surface Correspondences.}
One of the most widely used approaches in this topic is DensePose~\cite{neverova2018dense}, where classification and regression branches were trained to obtain per-pixel body part and UV estimates. The body parts constitute the I channel which takes one of $25$ values (including the background) and the UV estimates which are continuous numbers mapped to $[0, 255]$. However, its output is discretized resulting in seams between body parts. This problem is alleviated in Continuous Surface Embeddings (CSE)~\cite{neverova2020continuous}, which for each pixel learns a positional embedding of the corresponding vertex in the object mesh. In CSE correspondences are learned without being constrained on specific geometry types (\eg, humans), and show the effectiveness of their approach on other deformable object categories, like animal classes which was later extended by discovering correspondences between different object classes automatically~\cite{neverova2021discovering}.
HumanGPS~\cite{tan2021humangps} maps each pixel
to a feature space, where the feature distances reflect
geodesic distances among vertices of a 3D body model corresponding to every pixel. Similarly to CSE, for every image pixel they produce an embedding capable of differentiating visually similar parts and aligning different subjects into an unified feature space.
Zeng \etal~\cite{zeng20203d} introduced, a model-free 3D human mesh estimation framework, which explicitly establishes the dense correspondences between the mesh and the local image features in the UV space. They solve human body estimation problem relying on dense local features transferred to the UV space.
Getting enough labeled data (especially non-synthetic) to learn dense correspondences is a challenging task. SimPose~\cite{zhu2020simpose} proposed to alleviate the problem by using simulated multi-person datasets and a specific training strategy with multi-task objectives to learn dense UV coordinates. They obtain favourable results using only simulated human UV
labels. The intricacy of getting dense and accurately annotated
correspondences is further explored in UltraPose~\cite{yan2021ultrapose}. They provide a dense synthetic benchmark focusing on faces, containing around $1.3$ billion corresponding points as well as data generation system based on novel decoupling 3D model.

\noindent \textbf{Architecture Designs for Dense Correspondences.} There have been a couple of approaches in terms of network architectures to extract dense human correspondences. In DensePose~\cite{neverova2018dense} a Mask-RCNN~\cite{he2017mask} with Feature Pyramid Features~\cite{lin2017feature} is utilized to obtain accurate image features. 
SimPose~\cite{zhu2020simpose} opted for a ResNet-101 backbone trained with losses adjusted to each of their tasks (\eg, human pose, segmentations, normals, UVs). Another simple yet effective choice employed by HumanGPS~\cite{tan2021humangps} is an Encoder-Decoder architecture such as U-Net~\cite{ronneberger2015u}. 
Our investigation indicated that while one can achieve satisfactory results with the aforementioned approaches they are all unable to capture finer-level details in the depicted human as usually the extracted features are too coarse. To alleviate this we turned into transformer architectures due to their ability to learn these discriminative features necessary for either downstream computer vision tasks or reconstruction applications. 
Originating from Natural Language Processing, a Transformer architecture~\cite{vaswani2017attention} has shown its effectiveness within a wide range of Computer Vision tasks: image recognition and classification \cite{dosovitskiy2020image}, image retrieval \cite{el2021training}, image generation \cite{parmar2018image} and image captioning \cite{he2020image}. 
We capitalize upon prior work on vision transformers for dense prediction tasks~\cite{Ranftl2021} (\eg, depth estimation) and introduce a new architecture explicitly design for predicting dense surface correspondences for humans. 

\section{The Proposed Method: BodyMap}\label{sec:method}
The main goal of the proposed approach is to establish dense surface correspondence between a single RGB image and 3D body model. Our method takes as input a single RGB image, foreground mask and coarse correspondences retrieved using Continuous Surface Correspondences (CSE)~\cite{neverova2020continuous}. CSE serves as a sufficient initialization which our method refines by providing more accurate estimates for the areas covering loose clothes, hair, fingers, etc. Thus, BodyMap provides per-pixel estimates for the foreground image resulting in much more accurate representations and crisp silhouettes. The necessity of foreground mask stems not only from the foreground silhouette that we aim to complete with our estimates but also the image-level features that we prove to be essential in Section~\ref{sec:exp}.

\subsection{Continuous Correspondences}
Continuous correspondences have significant advantages over their discretized counterparts. First, a continuous representation provides no seams between body parts. Second, it is conceptually simpler as there is no need to explicitly encode and later predict the body part. The benefits of utilizing a continuous representation for surface correspondences have already been discussed in a few prior works~\cite{neverova2020continuous, tan2021humangps,zeng20203d}. We follow a similar direction with~\cite{neverova2020continuous}  and design a continuous UV map that is then warped to body models in different poses, providing ground truth correspondences for our approach. The color scheme for correspondences used in the paper is unique color-wise: we chose different colors for every vertex of the parametric body model. Given the colored 3D body model we transform its surface into a 4K UV map, which is then utilized during rendering over a body model in a determined pose and from a desired view point. In that way, we obtain ground truth for the synthetic data used for training.  

\subsection{Surface Embedding Transformers}
A classic architectural for a network predicting Dense Correspondences from an RGB image is an Encoder-Decoder (\eg, U-net). While simple convolutional backbones in the encoder can usually provide sufficient results, we observed that the right choice of the encoder architecture may significantly boost the whole pipeline performance. Compared to convolutions, transformer-based architectures do not suffer from limited receptive fields, resulting in more expressivity. Moreover, transformers avoid explicit downsampling of the input image embedding leading to more accurate and refined final representations.

As illustrated in Fig.~\ref{fig:hd-cse}, we build upon the work of Ranftl \etal~\cite{Ranftl2021} for monocular depth estimation and introduce a simple yet novel transformer-based architecture designed explicitly for the task of predicting dense surface correspondences of humans. We transform the RGB image and its CSE estimate into tokens by extracting non-overlapping patches and then linearly projecting resulting flattened representations. Similarly to text-transformers, we add a specific token to the set, that aggregates the global knowledge about an image. The image and CSE embeddings are supplemented with positional embeddings and fed to separate vision transformer backbones with separate weights to retrieve dense features for each input. Later we refer to these blocks as appearance and correspondence transformers (Fig.~\ref{fig:hd-cse}). Positional encoding in Visual Transformers is essential to capture sequence ordering of input tokens instead of transforming the image into "bag-of-patches" omitting its relative order and global spatial consistency.

The transformer outputs are fused forming an intermediate representation which is first resampled and then projected via residual convolutional units. It is then fed into the convolutional decoder where the representation is upsampled to generate a fine-grained correspondence prediction. Finally, the network outputs per-pixel RGB values that encode correspondences according to our coloring scheme discussed in the previous section.

\vspace{-0.1cm}
\subsection{Supervision in the Image Space}\label{ssec:super_image}
For each pixel $p$ in the foreground image, we predict 3-channeled (RGB) color $p' \in \mathbb{Z}^3$ which represents the correspondence (the colors in such a representation are unique which makes subsequent warping easy). Thus, we treat the whole problem as a multi-task classification problem where each task (predictions for the R, G and B channels) is trained with the same set of losses:

\noindent \textbf{Per-pixel classification loss} $L_{cls}$. For every color channel, we predict the per-pixel classification label $l \in [0, 255]$. 

BodyMap provides raw, unnormalized per-pixel scores for each of the classes in each of the three color channels and $L_{cls}$ measures the cross-entropy between the prediction and the ground truth. 
Since we noticed that it is quite challenging to predict correspondences of realistic gestures, we further define a loss weight for each pixel based on the body part segmentation. We set a higher weight for \textit{hands} and \textit{head} while a lower weight for the rest of the body to encourage fine-grained correspondence estimation.

\noindent \textbf{Silhouette loss} $L_{sil}$. We penalize the model for non-accurate silhouette predictions by calculating the IoU between the predicted and ground truth foreground masks.

\vspace{-0.1cm}
\subsection{Supervision in the 3D Geometry}\label{ssec:super_3d}

\noindent \textbf{Geodesic loss} $L_{geo}$. While per-pixel cross-entropy classification losses supervise our predictions in 2D image space, we expand our supervision scheme to 3D by utilising geodesic distances on the surface of the body model. Geodesic losses have been instrumental in the literature for enforcing supervisions in the 3D space. We design a loss that pushes features between non-matching pixels apart, depending on the geodesic distance. We calculate geodesic distances between vertices predicted with correspondences for every foreground pixel and their ground truth counterparts. Theoretically, such a supervision eliminates imperfection of the proposed coloring scheme for correspondences: distant vertices may have resembling colors (green head and shoulders, blue arm and right thigh). Thus, the geodesic loss provides extra knowledge about the 3D geometry comparing distances between predicted vertices vs. ground truth ones. \\
\begin{equation}\small
L_{geo}(I_{pred}, I_{gt}) = \\ 
\sum_{x} \mathcal{D}_g(V^{(I_{pred}(x))}, V^{(I_{gt}(x))}),
\end{equation}
where $V^{(I(x))}$ denotes the vertex corresponding to pixel location $x$ in the image $I$ and $\mathcal{D}_g(\cdot,\cdot)$ denotes the geodesic distance between two 3D points on the body surface.
\vspace{-0.1cm}
\subsection{Regularization and Final Loss}\label{ssec:super_reg}
\noindent \textbf{Consistency loss} $L_{con}$. We further add a regularization term to enforce the smoothness of the predictions in the neighboring regions. Specifically, we 
constrain the predictions from neighboring pixels to be geodesically close to each other, \ie, 
\begin{equation}\small
L_{con}(I_{pred}) = \sum_{p \in I} \log{\Bigg(1 +\exp\Big(\dfrac{\mathcal{D}_g(p_r, p)}{\sigma_{geo}} - \dfrac{\|p_r - p\|_1}{\sigma_{col}}\Big)\Bigg)},
\end{equation}
where $p_r$ is a randomly chosen pixel within the foreground silhouette, $\mathcal{D}_g(p_1,p_2)$ the geodesic distance between vertices corresponding to pixels $p_1$ and $p_2$, $\sigma_{geo}$ is the normalizing constant for geodesic distances (maximum possible distance between points in the body model), $\sigma_{col}$ the normalizing constant for RGB colors, respectively. On each iteration we calculate this loss $100$ times for different randomly chosen pixels later averaging the resulting values.

\noindent\textbf{Final loss}. The final loss is a weighted sum of all the terms:
\begin{equation}\small\begin{aligned}
L_{train} = &\lambda_{cls} L_{cls} + \lambda_{sil} L_{sil} + 
\lambda_{geo} L_{geo} +
\lambda_{con} L_{con},
\end{aligned}
\end{equation}
where a loss weight $\lambda$ corresponds to each loss term in order to balance them. 

\subsection{Training Details}\label{ssec:tr_details}
The BodyMap network is first trained on synthetic data to learn surface correspondences for every foreground pixel. Given an RGB image, we obtain foreground mask and CSE estimates which serve as an initialization for the correspondences. However, if we were to test this model directly on DensePose-COCO that comprises multiple people, heavy occlusions and low-resolution images, then the results would be unsatisfactory. The annotations provided in this dataset are sparse and noisy with \mytilde$100$ pixel-SMPL vertex correspondences for each person in the image. 
To bridge this domain gap, we fine-tune our model on the training set of DensePose-COCO but with a key change that ended up having a significant impact. Given an image from this dataset, we generate pseudo ground-truth estimates on the fly by extrapolating both the available ground-truth annotations but also the CSE initialization such that they cover the whole estimated silhouette of the human. In that way we can fine-tune our model on real-data with denser supervision and utilize losses in both 2D and 3D spaces. 

To further boost the generalization capabilities of \bodymap{}, we introduce several augmentations. First, we do specific crops in order to get upper-body samples. Second, we generate frames with multiple synthetic people in it to simulate crowds and diminish the gap between synthetic and real data. Third, we do a standard set of augmentations, like rotations, slight hue and saturation changes.

\section{Experiments}\label{sec:exp}

\noindent \textbf{Datasets}. Our proposed approach is trained mainly on synthetic data with the exception of the experiments reported on DensePose-COCO where we utilize the provided training set. We opted for the RenderPeople dataset~\cite{renderppl} which has been used extensively in the literature~\cite{alldieck2019tex2shape, bhatnagar2019mgn, huang2020arch, he2021archplus,lazova2019360, palafox2021spams, saito2020pifuhd, tan2021humangps, agora, 3Dhumantexturesynthesis,zins2021data} for various human reconstruction and generation tasks. We used $1000$ scans which are watertight meshes wearing a variety of garments and in some cases holding objects such as mugs or bags. Since the scans are static we wanted to introduce additional pose variations and as a result we performed non-rigid registration, rigged them for animation and used a motion collection that provides 3D human animations from which we collect a set of $2,446$ 3D animation sequences covering wide action categories of daily activities and sports. With a large set of scans and motions we randomly sample scan-motion pairs and render them with Blender Cycles from different views with uniform lighting to obtain the RGB sequences as well as the corresponding UV ground-truth. We perform a 90/10 train/test split based on identities. This large-scale dataset represents an effort to cover a wide range of motions, poses, and body shapes, captured from multiple views with people that can move towards the camera our even outside the frame and enables us to train our \bodymap{} network without making any explicit assumptions.

\begin{table}[ptb]
    \small
    \vspace{-2mm}
    \setlength{\tabcolsep}{0.6em}
    \hspace*{-0.4em}
    \resizebox{1.02\linewidth}{!}{
    \begin{tabular}{@{}lcccccc@{}}
    \toprule
    Method  & $AP$ & $AP_{50}$  & $AP_{75}$  & $AR$  & $AR_{50}$  & $AR_{75}$ \\ \midrule
    AMA-net~\cite{guo2019adaptive} & 64.1 & 91.4 & 72.9 & 71.6 & 94.7 & 79.8 \\
    DensePose~\cite{alp2018densepose} & 66.4 & 92.9 & 77.9 & 71.9 & 95.5 & 82.6 \\
    DensePose-DeepLab~\cite{alp2018densepose}  & 51.8 & 83.7  & 56.3 & 61.1 & 88.9 &  66.4 \\
    SimPose-Rendppl.~\cite{zhu2020simpose} & 57.3 & 88.4 & 67.3 & 66.4 & 95.1 & 77.8 \\
    SimPose-SMPL~\cite{zhu2020simpose} & 56.2 & 87.9 & 65.3 & 65.2 & 95.1 & 75.2 \\
    CSE~\cite{neverova2020continuous}  & 67.0 & 93.8 & 78.6 & 72.8 & 96.4 & 83.7\\
    CSE-DeepLab~\cite{neverova2020continuous}   & 68.0 & 94.1 & 80.0 & 74.3 & \textcolor{blue}{97.1} & 85.5 \\
    \hline
    BodyMap RGB-only  & \textcolor{blue}{71.0} & \textcolor{blue}{94.3} & \textcolor{blue}{83.3} & \textcolor{blue}{75.2} & 94.3 & \textcolor{blue}{86.1}  \\ 
    BodyMap   & \textbf{75.2} & \textbf{95.8} & \textbf{89.7} & \textbf{79.8} & \textbf{97.3} & \textbf{89.7} \\ \bottomrule
    \end{tabular}
    }
    \centering
    \beforetab
    \caption{\textbf{Average Precision (AP) and Recall (AR) on DensePose-COCO}. AP and AR are calculated at a number of GPS thresholds ranging from 0.5 to 0.95. Our methods surpasses the state-of-the art methods DensePose~\cite{neverova2018dense} and CSE~\cite{neverova2020continuous}}
    \aftertab
    \label{tab:COCO_3d}
\end{table}

At test-time \bodymap{} is evaluated quantitatively and qualitatively on both synthetic as well as real data ranging from COCO, fashion images (DeepFashion~\cite{liuLQWTcvpr16DeepFashion}) as well as a few 3dMD scans of real people captured with a full-body scanner. We used this solely for testing since we wanted to evaluate to what extent our approach can handle the domain gap between synthetic and real data. These real scans do not include any objects but are noisier with complex facial expressions and enable us to stress-test whether our approach can handle such complex inputs. 

\noindent \textbf{Baselines and Metrics}. We consider two different ways of measuring the quality of correspondences evaluating both in 2D image space by comparing RGB values of the corresponding pixels and in 3D space by measuring geodesic distances between predicted and ground truth vertices.

First, we calculate the accuracy of predictions in the 2D image space by calculating the percentage of pixels colored correctly within a specified threshold. Second, following the evaluation scheme of DensePose that is used widely in the literature~\cite{neverova2018dense, neverova2020continuous, zhu2020simpose} we measure average precision and recall over GPS scores. Geodesic point similarity (GPS) score is a correspondence matching score:

\begin{equation}\small
    \text{\textit{GPS}}_j = \dfrac{1}{|P_j|} \sum\nolimits_{p \in P_j} \exp{\dfrac{-g(i_p, \hat{i_p})^2}{2 \kappa^2}},
\end{equation}
where $P_j$ is the set of points annotated on person instance $j$, $i_p$ is the vertex estimated by a model at
point $p$, $\hat{i_p}$ is the ground-truth vertex $p$, and $\kappa$ is a normalizing parameter.
We calculate Average Precision (AP) and Average Recall (AR) metrics considering a vertex prediction as correct if the GPS score is higher than a threshold. Following the evaluation scheme established by prior work~\cite{neverova2018dense, neverova2020continuous}, GPS thresholds are ranging from 0.5 to 0.95.

Additionally to metrics in 2D and 3D spaces, we evaluate the consistency over time of our predictions in order to estimate quantitatively the amount of flickering. We calculate percentage of positive correspondence matches between frames of the same video for visible vertices.

\begin{table}[t]
    \small
    \setlength{\tabcolsep}{0.4em}
    \hspace*{-0.4em}
    \resizebox{\linewidth}{!}{
    \begin{tabular}{@{}lcccccc@{}}
    \toprule
     & \multicolumn{3}{c}{Synthetic Dataset} & \multicolumn{3}{c}{DensePose-COCO} \\ \cmidrule(l){2-4} \cmidrule(l){5-7} 
    Error Window (px)                   & 5           & 10         & 20         & 5          & 10        & 20        \\ \midrule
    DensePose~\cite{alp2018densepose}   & 25.93       & 46.10      & 69.91      & 49.23      & 55.75     & 59.71     \\
    CSE~\cite{neverova2020continuous}   & 44.52       & 67.51      & 75.13      & 58.10      & 60.34     & 64.14     \\
    \hline
    BodyMap RGB-only                       & 66.15       & 73.81      & 79.80      & 61.18      & 65.32     & 68.52     \\ 
    BodyMap                       & \textbf{71.12}       & \textbf{79.73}      & \textbf{96.92}      & \textbf{65.34 }     & \textbf{68.22}     & \textbf{73.88}     \\ \bottomrule
    \end{tabular}
    }
    \centering
    \beforetab
    \caption{\textbf{Accuracy in 2D space}. We show the percentage of pixels correctly matched within the established error window on synthetic dataset and DensePose-COCO. Our methods surpasses the state-of-the art methods DensePose and CSE.}
    \aftertab
    \label{tab:2d_metrics}
\end{table}

Using the aforementioned metrics we compare \bodymap{} quantitatively to the previous works: DensePose~\cite{neverova2018dense}, CSE~\cite{neverova2020continuous}, SimPose~\cite{zhu2020simpose} as well as several other baselines. However, calculating AP and AR metrics for HumanGPS is not possible due to the fact, that HumanGPS predicts only embeddings for every foreground pixel, that provide no information on UV coordinates or corresponding to pixels SMPL vertices. In their approach warping and appearance swapping is done by nearest neighbors search over embeddings without going to 3D body model space. Thus, we compare with HumanGPS only qualitatively and using temporal consistency metrics.

\subsection{Quantitative results}
In Tables~\ref{tab:2d_metrics} and \ref{tab:TrueTony_3d} we provide a quantitative comparison between BodyMap, DensePose, CSE and HumanGPS on the test set of the aforementioned synthetic dataset. In Tables~\ref{tab:COCO_3d} and~\ref{tab:2d_metrics} we do the same on DensePose-COCO dataset but also provide additional comparisons with prior work. Opposite to the synthetic dataset, for which we have ground truth correspondences for every foreground pixel, for DensePose-COCO we rely only on the available annotated points to calculate the metrics. BodyMap shows a substantial improvement over prior work across all the metrics for both our synthetic and DensePose-COCO datasets. The reasons behind this improvement stem from: i) specifically designed architecture that separates and takes the best out of RGB and CSE inputs; ii) training on well-designed and rendered synthetic data and later fine-tuning on specifically adapted DensePose-COCO with the additional tricks discussed in Sec.~\ref{ssec:tr_details}, which helps to bridge the synth2real domain gap; iii) the proposed training scheme that includes supervision both in the image space with per-pixel classification losses as well as the 3D space with geodesic losses.

\begin{table}[ptb]
    \small
    \setlength{\tabcolsep}{0.2em}
    \resizebox{\linewidth}{!}{
    \begin{tabular}{@{}lcccccccccc@{}}
    \toprule
    Method                   & $AP$           & $AP_{50}$         & $AP_{75}$         & $AP_M$          & $AP_L$        & $AR$           & $AR_{50}$         & $AR_{75}$         & $AR_M$          & $AR_L$         \\ \midrule
    DP-DL~\cite{alp2018densepose}   & 55.3 & 85.6  & 60.1 & 48.3 & 58.2 & 66.8 & 90.1 &  68.2 & 50.1 & 66.1      \\
    CSE-DL~\cite{neverova2020continuous}   & 72.8 & 95.7 & 84.2 & 65.7 & 73.1 & 78.2 & 97.3 & 87.5 & 67.2 & 78.0       \\
    \hline
    BodyMap RGB-only  & 75.3 & 96.1 & 89.2 & 69.3 & 75.2 & 81.2 & 97.4 & 89.2 & 70.3 & 80.2      \\ 
    BodyMap   & \textbf{79.5} & \textbf{97.8} & \textbf{90.5} & \textbf{72.3} & \textbf{79.4} & \textbf{85.3} & \textbf{98.1} & \textbf{92.5} & \textbf{73.4} & \textbf{84.5}      \\ \bottomrule
    \end{tabular}
    }
    \centering
    \beforetab
    \caption{\textbf{Average Precision (AP) and Recall (AR) over GPS scores in 3D space}. We calculate AP and AR at GPS thresholds ranging from 0.5 to 0.95 on our synthetic dataset. Our method clearly outperforms DensePose-DeepLab and CSE-DeepLab.}
    \aftertab
    \label{tab:TrueTony_3d}
\end{table}

\noindent \textbf{Temporal Consistency}. In Table~\ref{tab:time_consistency} we test how temporally consistent the dense correspondences of different methods are. We were motivated to run this experiment by observing how jittery DensePose predictions can be on videos. In terms of metrics, we estimate the percentage of positive correspondence matches between the current frame and a frame in the future with an interval in \({1,12,120}\) on a synthetic sequence consisting of $18,000$ frames. \bodymap{} outperforms prior work by a large margin and establishes accurate correspondences even if the time interval between the 2 frames is substantial. In supplementary we provide demo video showing consistency over time of our results.

\begin{table}[t]
    \centering
    \setlength{\tabcolsep}{1.8em}
    \resizebox{\linewidth}{!}{
    \begin{tabular}{@{}lccc@{}}
        \toprule
         Frame Interval & 1 & 12 & 120 \\
         \midrule
         DensePose~\cite{neverova2018dense} & 77.79 &	40.86 &	16.32  \\
         CSE~\cite{neverova2020continuous} &    85.55	& 55.85	& 18.93 \\
         HumanGPS~\cite{tan2021humangps} &  86.42 & 65.19 & 36.17 \\ 
         \midrule
         BodyMap~ &    \textbf{88.70}       & \textbf{74.01}      & \textbf{46.11} \\ 
         \bottomrule
    \end{tabular}
    }
    \beforetab
    \caption{\textbf{Temporal consistency}. We estimate the percentage of positive correspondence matches between frames with a different interval on a synthetic sequence of $18,000$ frames.}
    \aftertab\vspace{-1mm}
    \label{tab:time_consistency}
\end{table}

\subsection{Ablation studies}\label{ssec:ablations}
\noindent \textbf{Different Architectures}. We experiment with different backbones starting from a simple UNet with skip-connections and then progressing to more complex transformer-based solutions. In Table~\ref{tab:ablation_architectures} we provide a comparison in terms of accuracy in the 2D space across all the architectures. An interesting finding is that a simple UNet architecture can get satisfactory results when trained with all the proposed supervisions described in Sec.~\ref{ssec:super_image} and~\ref{ssec:super_3d}. However, our proposed Vision Transformer (ViT) is capable of learning more accurate correspondences in challenging areas like neck, armpits, fingers and hair, making the predicted silhouette clear-cut and crisp. These differences are mostly visible in hard DensePose-COCO examples (with multiple people and occlusions), while on simple synthetic data cases UNet is performing nearly as good as ViT.

We further experiment with the network design, feeding only RGB inputs to the net and omitting the Correspondence Transformer. While RGB-only method performs comparatively worse, it still outperforms existing approaches, e.g.DensePose, CSE or HumanGPS (Tables~\ref{tab:COCO_3d}, \ref{tab:2d_metrics}).

\begin{table}[t]
    \vspace{2mm}
    \centering
    \small
    \setlength{\tabcolsep}{0.4em}
    \begin{tabular}{@{}llcccccc@{}}
    \toprule
    &        & \multicolumn{3}{c}{Synthetic Dataset} & \multicolumn{3}{c}{DensePose-COCO} \\ 
    \cmidrule(l){3-5} \cmidrule(l){6-8} \multicolumn{2}{c}{Error Window (px)} & 5 & 10 & 20 & 5 & 10 & 20 \\ \midrule
    \multirow{4}{*}{\begin{tabular}[c]{@{}l@{}}BodyMap\\ (ours)\end{tabular}} & ResNet & 45.12       & 60.82     & 79.12      & 30.41     & 55.67     & 61.15        \\
    & EffNet & 51.22       & 65.77     & 82.19      & 40.25     & 61.17     & 70.22        \\
    & U-Net  & 68.42       & 75.13      & 94.19      & 60.82     & 65.74     & 70.12 \\
    & ViT     & \textbf{71.12}       & \textbf{79.73}      & \textbf{96.92}      & \textbf{65.34}      & \textbf{68.22}     & \textbf{73.88}        \\
    \bottomrule
    \end{tabular}
    \beforetab
    \caption{\textbf{Different network backbones}: Ablation study}
    \aftertab
    \label{tab:ablation_architectures}
\end{table}

\noindent \textbf{Different Losses}. We also investigate the impact of the proposed losses in Table~\ref{tab:ablation_losses}. While the best score is achieved with the whole set of proposed losses,  per-pixel cross-entropy classification losses for color channels contribute the most. Silhouette loss makes the edges of final prediction more accurate and extra supervision in hands and head regions improves correspondences in these areas. Geodesic losses give tangible improvement only on DensePose-COCO, indicating that simple synthetic one-person-per-frame cases can be handled sufficiently with only image-space supervision. Thus, the model can learn fine-grained body model details even with the first two losses (both supervising in the 2D image space). However, more complicated cases including several people in one frame and significant occlusions require extra supervision in 3D space to obtain satisfactory results.

\begin{table}[t]
    \centering
    \setlength{\tabcolsep}{0.2em}
    \resizebox{\linewidth}{!}{
    \begin{tabular}{@{}ccccccc@{}}
    \toprule
          &  \multicolumn{3}{c}{Synthetic Dataset} & \multicolumn{3}{c}{DensePose-COCO} \\
         \cmidrule(l){2-4} \cmidrule(l){5-7}
         \diagbox{Losses}{Error Window} & 5 & 10 & 20 & 5 & 10 & 20 \\
         \midrule
         \(L_{cls}\) &         65.16       & 71.52     & 85.12      & 49.37      & 55.81     & 59.14  \\
         \(L_{cls}+L_{sil}\) &    69.18       & 75.32     & 92.31      & 54.12      & 60.22     & 62.17 \\
         \(L_{cls}+L_{sil}+L_{geo}\) &    70.23       & 78.71     & 95.80      & 61.83      & 64.32     & 68.17 \\
         \(L_{cls}+L_{sil}+L_{geo} + L_{con}\) &    \textbf{71.12}       & \textbf{79.73}      & \textbf{96.92}      & \textbf{65.34}      & \textbf{68.22}     & \textbf{73.88} \\
         \bottomrule
    \end{tabular}
    }
    \beforetab
    \caption{\textbf{Ablation study on the impact of different losses} in the accuracy in 2D space (the percentage of pixels colored correctly within the established error window)}
    \aftertab
    \label{tab:ablation_losses}
\end{table}

\noindent \textbf{Different Fine-tuning Schemes}: We experimented with two ways of fine-tuning on real data: (1) using only available sparse annotations (sparse fine-tuning); (2) using the generated dense pseudo ground-truth estimates described in Sec.~\ref{ssec:tr_details} (dense fine-tuning). We observed that densifying ground-truth on the fly results in superior performance compared to either no fine-tuning or relying solely on sparse annotations. More results are shown in the supplementary. 

\noindent\textbf{Model Complexity}: Inference of our model takes \mytilde$0.1$ seconds on a single Tesla V100-SXM2 for a $1024 \times 1024$ image. The model has ~\mytilde$600M$ trainable parameters. 

\vspace{-0.07cm}
\subsection{Qualitative Results}
\vspace{-0.08cm}

\begin{figure}[ptb]
    \vspace{-2mm}
    \centering
    \includegraphics[width=0.9\linewidth]{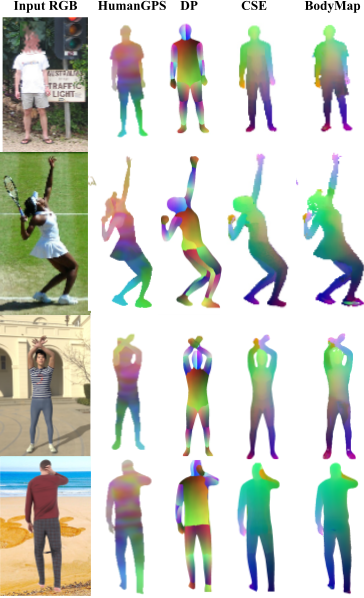}
    \vspace{-0.2cm}
    \caption{Qualitative comparison with competitors on DensePose-COCO, our synthetic dataset and RenderPeople.}
    \vspace{-0.5cm}
    \label{fig:comparison}
\end{figure}

In Figures~\ref{fig:teaser}, \ref{fig:comparison} we show correspondences for a few images from DeepFashion~\cite{liuLQWTcvpr16DeepFashion} which has lower quality inputs, RenderPeople, DensePose-COCO and finally images from real-people scans captured with a 3dMD system. The silhouettes of the inputs are well covered with our estimates, the hands and fingers are accurately captured and the face is well aligned. Loose clothes, even complicated cases like a long robe in the are well-handled.

In Figure~\ref{fig:comparison} we show qualitative comparisons between BodyMap and competitors: HumanGPS, DensePose and CSE on several examples from DensePose-COCO, RenderPeople and our synthetic dataset. While DensePose and CSE predictions are smooth and consistent, they do not cover the whole silhouette, totally omitting hair and loose clothes. HumanGPS handles silhouettes better, but still struggles with accurate correspondences in challenging scenarios with occlusions or produces blurry patches for back views (Line~$4$ in Figure~\ref{fig:comparison}). We also show in the supplementary, that HumanGPS predictions are not always temporally consistent, jumbling correspondences for right and left arms and legs while the person is rotating. 

\vspace{-0.08cm}
\subsection{Applications \& Discussions}
\vspace{-0.1cm}

\begin{figure}[ptb]
    \vspace{-2mm}
    \centering
    \includegraphics[width=\linewidth]{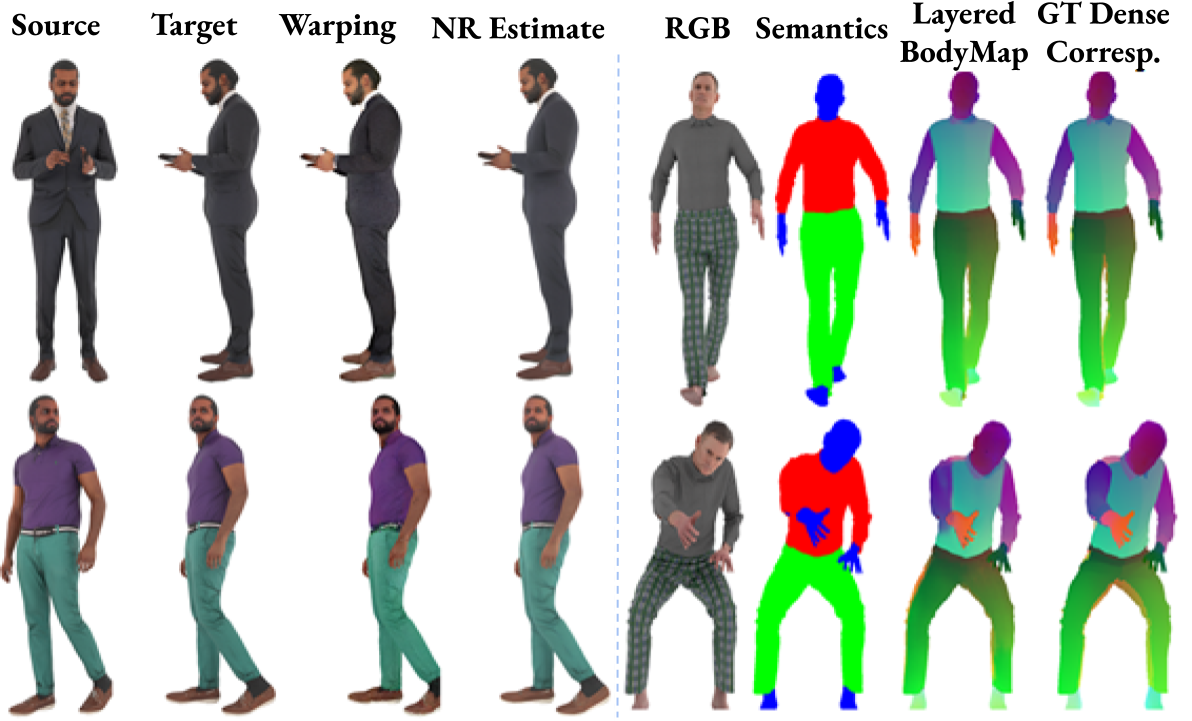}
    \beforefigcaption
    \caption{\textbf{Applications}. Neural re-rendering (left) and predicting layered correspondences for clothed humans (right).}
    \afterfigcaption
    \label{fig:applications}
\end{figure}

\noindent\textbf{Neural Re-rendering}. One possible application is re-rendering people from the source frame from another view point and/or in another pose. 
We introduce a model for neural re-rendering which aims at learning a function that given the complete texture map and the estimated BodyMap correspondences generates a photorealistic render in the image space. Before neural re-rendering it is needed to obtain a complete texture map, which we do in the following way. 
Given a source and a target view of a person we utilize the predicted BodyMap estimates and defined a warping function \(W\) that outputs high-quality neural re-renders at the target viewpoint. We represent \(W\) with a neural network that i) warps the input source RGB image to the UV space to obtain a partial texture, ii) learns to complete it to obtain a full texture estimate, and iii) warps it back to the image space using BodyMap and then uses a neural renderer that generates the final output render. Given a source and a target image our neural renderer generates overall higher-fidelity details than prior work as seen in Fig.~\ref{fig:applications}(left), also in the face and hand regions, and does not suffer from color bleeding. In the supplementary material we describe in detail this application along with an architecture figure and also present an application to cloth swapping and motion retargeting.

\noindent\textbf{Layered Dense Correspondences}.
In all prior work dense human correspondences are estimated only for the body surface. That is because a body template (\eg, SMPL~\cite{loper2015smpl}) with UV information is available and sparse annotations for COCO exist to accomplish this task. However, when dealing with clothed humans (and especially in loose garments) estimating body correspondences in a single-layer as DensePose or our proposed \bodymap{} does, can be a challenging task. However, fine-grained clothes details like wrinkles and textile folds can be represented better with decoupling body and clothes correspondences to separate representations. In a first attempt to do so we present an application with a slight BodyMap variation predicting three separate representations for the unclothed body, upper clothes and lower clothes. We named this variation \emph{Layered-BodyMap}. The architecture remains the same besides the three output heads instead of one. 
To generate ground truth data for such a task, we run cloth simulation for the two garments given various walking and hand-movement motions resulting in $12$ sequences of people wearing 3D clothes from our collection. Opposite to BodyMap, where we use RGB together with CSE initialization as input, here we do not have any initialization for the clothes correspondences, and as a result we feed this network with RGB-only inputs, but condition the estimates on semantic segmentation masks. The predicted \emph{layered} correspondences are accurate and cover the whole silhouette (Fig.~\ref{fig:applications} (right)) which is a promising result that we believe future work will improve upon as more 3D garment libraries become available~\cite{santesteban2021self, tiwari2020sizer, zhu2020deep,bertiche2020cloth3d}. 

\begin{figure}[ptb]
    \vspace{-2mm}
    \centering
    \includegraphics[width=0.9\linewidth]{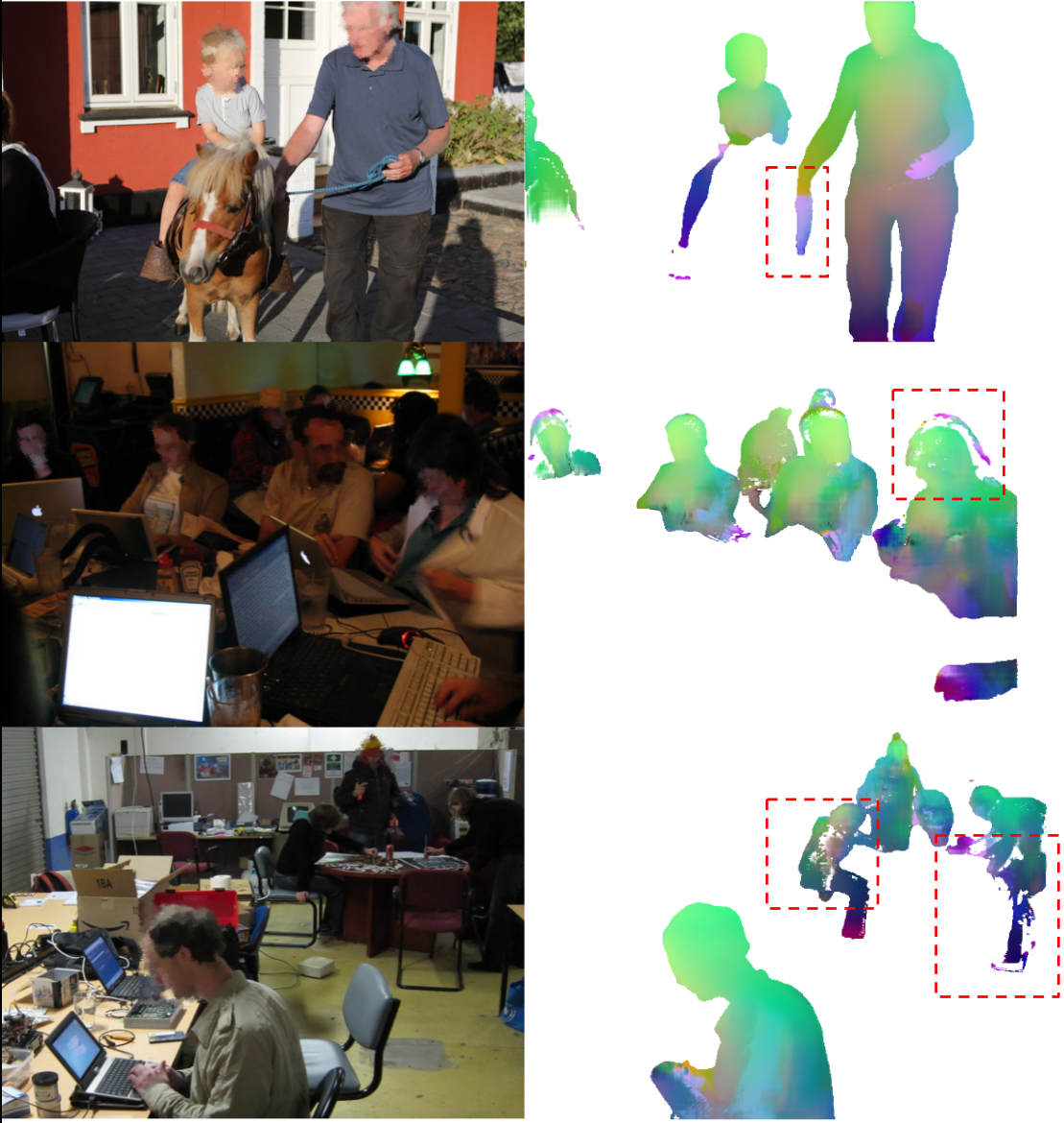}
    \beforefigcaption
    \vspace{0.2cm}
    \caption{\textbf{Failure cases}. Most failure cases happen for low resolution images with occlusions and/or bad lightning.}
    \afterfigcaption
    \vspace{-0.1cm}
    \label{fig:failure_cases}
\end{figure}

\noindent\textbf{Limitations}. Our approach relies on foreground human segmentation which makes it susceptible to the performance of that step. We tested different segmentation and matting approaches,~\cite{kirillov2020pointrend, MODNet, BMSengupta20, mmsegm}, and opted for MMSegmentation due to its ability to preserve fine details like fingers and hairstyles. BodyMap was trained on high-resolution mostly full-body images which makes it susceptible to low-res inputs or when only the bottom of the body is visible. This is partly solved by imposing heavy augmentations but occlusions from objects remain a challenge. Moreover, due to the nature of the task most of the training data is synthetic which makes inference on real data challenging. We address that with the fine-tuning scheme described in Sec.~\ref{ssec:ablations}. We show some failure cases in Figure~\ref{fig:failure_cases}, which mostly happen due to bad lightning or severe occlusions and provide additional examples in the supplementary.

\section{Conclusion}\label{sec:conc}
We present a novel framework for establishing accurate dense correspondences between an image and the surface of a 3D clothed human.
Our key contribution, BodyMap, is a transformer-based architecture that when trained with 2D and 3D supervisions significantly outperforms prior work. BodyMap addresses key limitations of current approaches, such as inability to handle loose clothes, body and garments being represented as a single surface, non-continuity of the correspondences for different body parts. We outperformed prior work by a large margin on synthetic as well as DensePose-COCO datasets and investigated the impact of each of our design selections. Finally, we provided examples of applications such as re-rendering in different poses and extend BodyMap to clothed humans with multiple layers of geometry with promising results.

\noindent \textbf{Acknowledgments.} We thank Tuur Stuyck for his help to run cloth simulation and Vasil Khalidov for his help with running the code of the Continuous Surface Embeddings paper. 
{\small
\bibliographystyle{ieee_fullname}
\bibliography{Refs}
}

\clearpage

\section*{Supplementary Material}

In this supplementary material we provide information regarding the broader impact of our method (Sec.~\ref{sec:impact}) additional details regarding fine-tuning our proposed BodyMap on real data (Sec.~\ref{finetuning}) and an in-depth discussion on applications to neural re-rendering and cloth swapping with several qualitative examples (Sec.~\ref{appls}). Additional qualitative evaluations and results are shown in the supplemental video.

\begin{figure}[h]
	\centering
	\includegraphics[width=\linewidth]{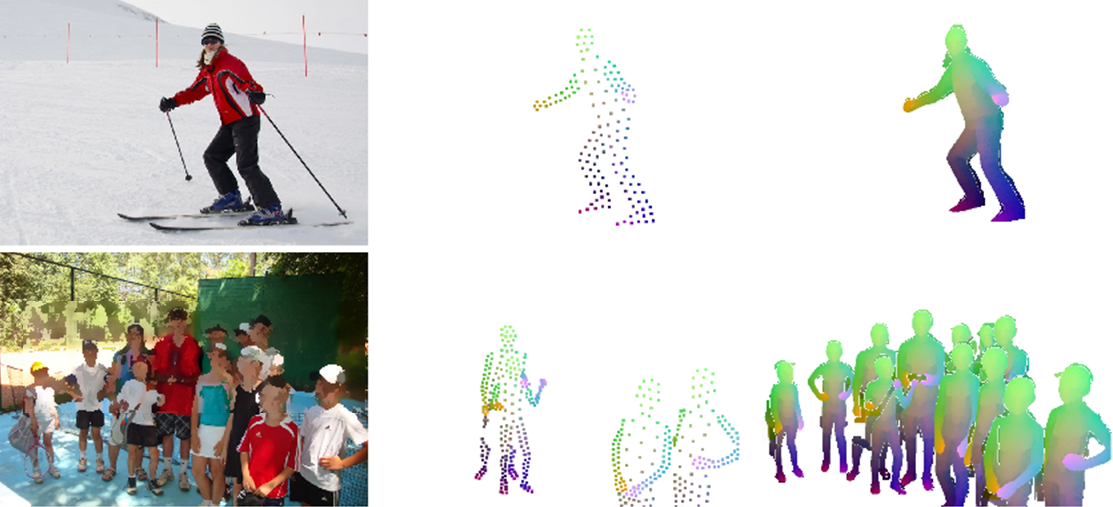}
	\caption{\textbf{Sparse and dense ground truth for DensePose-COCO dataset}. Sparse ground truth: annotation available in the dataset. Dense ground truth: pseudo ground-truth estimates are generated on the fly by extrapolating both the available ground-truth annotations but also the CSE initialization such that they cover the whole estimated silhouette of the human}
	\label{fig:dense_sparse_gt}
\end{figure}

\section{Broader Impact}\label{sec:impact}
The positive impact of our technology is further discussed in the applications section, including novel view synthesis and appearance swapping. While every human-related technology may raise concerns of fraudulent activities, we should note that our approach does not reconstruct facial or any other features used for personality identification. Thus, it is not possible to identify a person using our method, which makes our technology safe.

\section{Fine-tuning BodyMap on real data}
\label{finetuning}
While BodyMap is trained with mostly synthetic data, we further fine-tune it with DensePose-COCO dataset. Available DensePose-COCO annotations are extremely sparse (around $100$ annotated pixels per person, also annotations are not present for small people in the background). Thus, we leverage a heuristically made dense pseudo ground truth generation scheme. Given an image from the dataset, we generate pseudo ground-truth estimates on the fly by extrapolating both the available ground-truth annotations but also the CSE initialization such that they cover the whole estimated silhouette of the human (Figure~\ref{fig:dense_sparse_gt}). In that way we can fine-tune our model on real-data with denser supervision and utilize losses in both 2D and 3D spaces.

\begin{table}[t]
	\centering
	\small
	\setlength{\tabcolsep}{0.2em}
	\begin{tabular}{@{}lccccc@{}}
		\toprule
		\diagbox{Method}{Window (px)} & 5 & 10 & 20 \\
		\midrule
		DensePose~\cite{neverova2018dense} &      49.23 & 55.75 & 59.71 \\
		CSE~\cite{neverova2020continuous} &     58.10 & 60.34 & 64.14 \\
		\hline
		BodyMap: no ft & 40.51 & 52.18 & 55.22 \\
		BodyMap: ft 1 & 56.12 & 60.02 & 63.15 \\
		BodyMap: ft 2 & \textbf{65.34} & \textbf{68.22} & \textbf{73.88} \\
		\bottomrule
	\end{tabular}
	\beforetab
	\caption{\textbf{Ablation study on the effectiveness of finetuning: metrics in 2D image space.} We illustrate the accuracy in 2D space on DensePose-COCO (the percentage of pixels correctly matched within the established error window) before fine-tuning (\textbf{no ft}), and after fine-tuning with: (i) sparse annotations with ~100 labeled points per person (\textbf{ft 1}), and (ii) dense annotations, obtained heuristically by interpolating annotations within silhouettes \textbf{ft 2}). We include the performance of DensePose and CSE for comparison.}
	\aftertab
	\label{tab:COCO_finetuning_2D}
\end{table}

We experimented with two ways of fine-tuning on real data: (1) using only available sparse annotations (sparse fine-tuning); (2) using the generated dense pseudo ground-truth estimates (dense fine-tuning). The results of the both schemes are presented in Tables \ref{tab:COCO_finetuning_2D} and \ref{tab:COCO_finetuning_3D}, showing that dense fine-tuning allows to get the best results beating competitive methods such as HumanGPS, DensePose and CSE.

\begin{figure*}[t]
	\centering
	\includegraphics[width=0.95\linewidth]{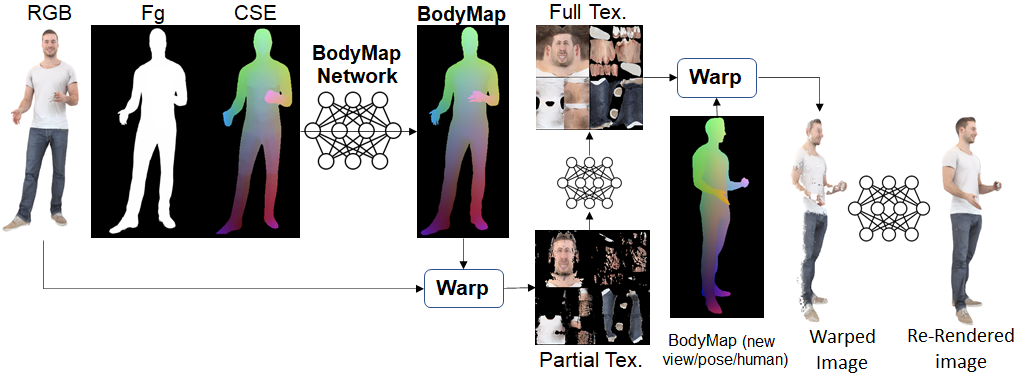}
	\caption{\textbf{Novel view rendering with BodyMap}. Given an RGB image we first obtain its CSE~\cite{neverova2020continuous} estimates and feed both to BodyMap network to get refined per-pixel predictions covering the whole silhouette. With such correspondences we can obtain partial texture maps that are very well aligned with the complete texture map which are then utilized to train a texture completion network to fill in the missing information. At last, we use the BodyMap as a mapping function back to the image space where we train a neural re-rendering framework that generates the photorealistic renders.}
	\label{fig:nn}
\end{figure*}

\begin{table}[ptb]
	\small
	\setlength{\tabcolsep}{0.2em}
	\resizebox{\linewidth}{!}{
		\begin{tabular}{@{}lcccccccccc@{}}
			\toprule
			Method                   & $AP$           & $AP_{50}$         & $AP_{75}$         & $AP_M$          & $AP_L$        & $AR$           & $AR_{50}$         & $AR_{75}$         & $AR_M$          & $AR_L$         \\ \midrule
			DP~\cite{alp2018densepose}   & 55.3 & 85.6  & 60.1 & 48.3 & 58.2 & 66.8 & 90.1 &  68.2 & 50.1 & 66.1      \\
			CSE~\cite{neverova2020continuous}   & 72.8 & 95.7 & 84.2 & 65.7 & 73.1 & 78.2 & 97.3 & 87.5 & 67.2 & 78.0       \\
			\hline
			BodyMap: no ft & 50.1 & 82.3 & 58.7 & 45.7 & 57.5 & 65.1 & 88.1 & 65.4 & 49.2 & 64.5      \\ 
			BodyMap: ft 1  & 70.3 & 94.2 & 83.1 & 63.2 & 72.1 & 77.8 & 96.5 & 85.1 & 65.9 & 76.1      \\ 
			BodyMap: ft 2   & \textbf{79.5} & \textbf{97.8} & \textbf{90.5} & \textbf{72.3} & \textbf{79.4} & \textbf{85.3} & \textbf{98.1} & \textbf{92.5} & \textbf{73.4} & \textbf{84.5}      \\ \bottomrule
		\end{tabular}
	}
	\centering
	\beforetab
	\caption{\textbf{Ablative studies on the effectiveness of finetuning: metrics in 3D space.} We illustrate Average Precision (AP) and Average Recall (AR) over GPS scores on DensePose-COCO dataset before fine-tuning (\textbf{no ft}), and after fine-tuning with: (i) sparse annotations with ~100 labeled points per person (\textbf{ft 1}), and (ii) dense annotations, received heuristically by interpolating annotations within silhouettes \textbf{ft 2}). We also include the performance of DensePose and CSE for comparison.}
	\aftertab
	\label{tab:COCO_finetuning_3D}
\end{table}

\section{Applications}\label{appls}
\subsection{Rendering people in arbitrary poses/views}
\label{NN}

One of the straightforward applications of our proposed BodyMap framework is re-rendering people in novel poses/views. Previous attempts to do so include Texformer~\cite{xu20213d}~---~Transformer-based framework for 3D human texture estimation from a single image.  They utilize pre-computed color encoding of the UV space obtained by mapping the 3D
coordinates of a standard human body mesh to the UV space as a Query, feeding it to the Transformer. We also rely on specifically designed Transformer-based architecture to retrieve dense correspondences that can lately be used as a mapping function between 2D image space and 3D space.
While we can do novel view rendering without any extra efforts simply warping the texture using BodyMap estimates as a mapping function, using additional neural renderer significantly increases the quality of the final render. Next we describe in the detail the proposed pipeline for generating novel views consisting of two main steps: texture completion network and neural rendering. The pipeline is also depicted in Figure~\ref{fig:nn}.

\begin{figure*}[t]
	\centering
	\includegraphics[width=\linewidth]{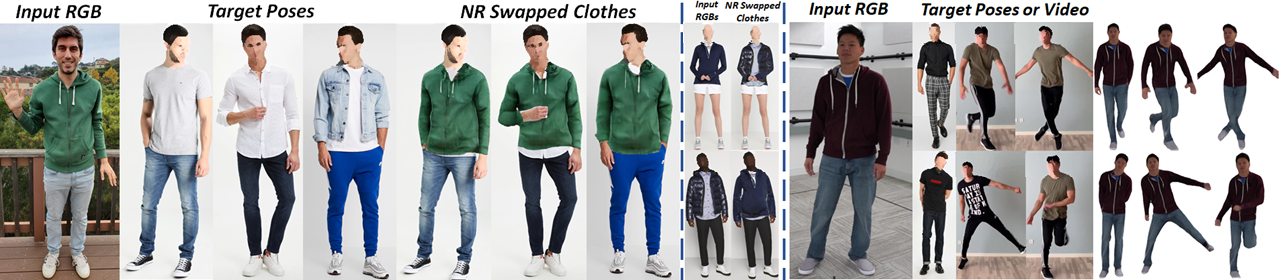}
	\caption{\textbf{Applications on virtual dressing and motion imitation:} given a single RGB image and a variety of target poses our method can dress the target subjects with the input hoodie at different poses with the hands being potentially occluded while preserving fine-level details. Our approach also generates photorealistic renders of the input human given target images of other people with different poses and viewpoints. Our method has not been trained specifically for any of these tasks but it can still generalize and produce crisp results.}
	\label{fig:appearance_swap}
\end{figure*}

\subsubsection{Texture Completion Network} 
Given the BodyMap estimates and the foreground RGB image we define a warping function that maps each foreground pixel of the image space to the UV-map space using BodyMap estimates as the mapping function. Note that in the UV map space, every point on the mesh surface of a human body template is represented by its coordinates on this UV map. We then train a texture completion neural network that takes as input partial textures at $1024\times1024$ resolution and completes the missing information by producing the full texture. The fact that partial texture is so well aligned on the UV map with full texture enables us to utilize a U-Net-like architecture since the skip connections can transfer the aligned input from the encoder to the decoder layers without adding an additional overheard to the decoder. A challenge that arises when dealing with high-resolution inputs is that only a few samples can fit into the GPU memory during training. Instance normalization blocks have widely been used in such cases to avoid collecting batch statistics that can be inaccurate due to the small sample size~\cite{alldieck2019tex2shape, lazova2019360}, but our experimental investigation indicated that instance normalization produces completed textures with distorted colors in non-visible regions. To overcome this challenge we propose to utilize synchronized batch normalization which differs from previous methods in the way the statistics are computed over all training samples distributed on multiple devices. This enables us to learn more accurate batch statistics that can then be used at test-time with traditional batch normalization blocks. Thus, given pairs of partial and full textures ($\{P_T, F_T\} \sim T$)
coming from the data distribution we train the texture completion network with the following losses:
\begin{itemize}
	\item Hinge version~\cite{lim2017geometric, zhang2019self} of the \textbf{adversarial loss}, along with a multi-scale discriminator as used in Pix2PixHD~\cite{isola2017image}:
	
	\begin{equation}
		L_G = - \mathbb{E}_{P_T \sim p_{P_T}, F_T \sim T} D(G(P_T), F_T)
	\end{equation}
	
	\begin{equation}
		\begin{multlined}
			L_D = -\mathbb{E}_{\{P_T, F_T\} \sim T} [0, -1 + D(P_T, F_T)] \\
			- \mathbb{E}_{P_T \sim p_{P_T}, F_T \sim T} [0, -1 - D(G(P_T), F_T)]
		\end{multlined}
	\end{equation}
	
	\item \textbf{Perceptual loss.} We utilize a pre-trained VGG~\cite{simonyan2014very} network and compute the $L_1$ loss between the completed texture estimate and the ground-truth texture map at the activations of five different layers of the network. Perceptual similarity losses help the network to generate fine-level details which are common in the textures of clothed humans.
	\item \textbf{Total variation loss.} We add a total variation (TV) loss~\cite{mahendran2015understanding} with a small weight in order to encourage spatial smoothness in the generated textures and remove some artifacts that are quite common in image-to-image translation networks. The TV loss is formulated as follows:
	\begin{equation}
		\begin{multlined}
			L_{TV} = \sum_{u} \sum_{v} |F_T(u, v+1) - F_T(u,v)| \\
			+ |F_T(u+1, v) - F_T(u,v)|
		\end{multlined}
	\end{equation}
\end{itemize}

\begin{figure*}[t]
	\centering
	\includegraphics[width=0.9\linewidth]{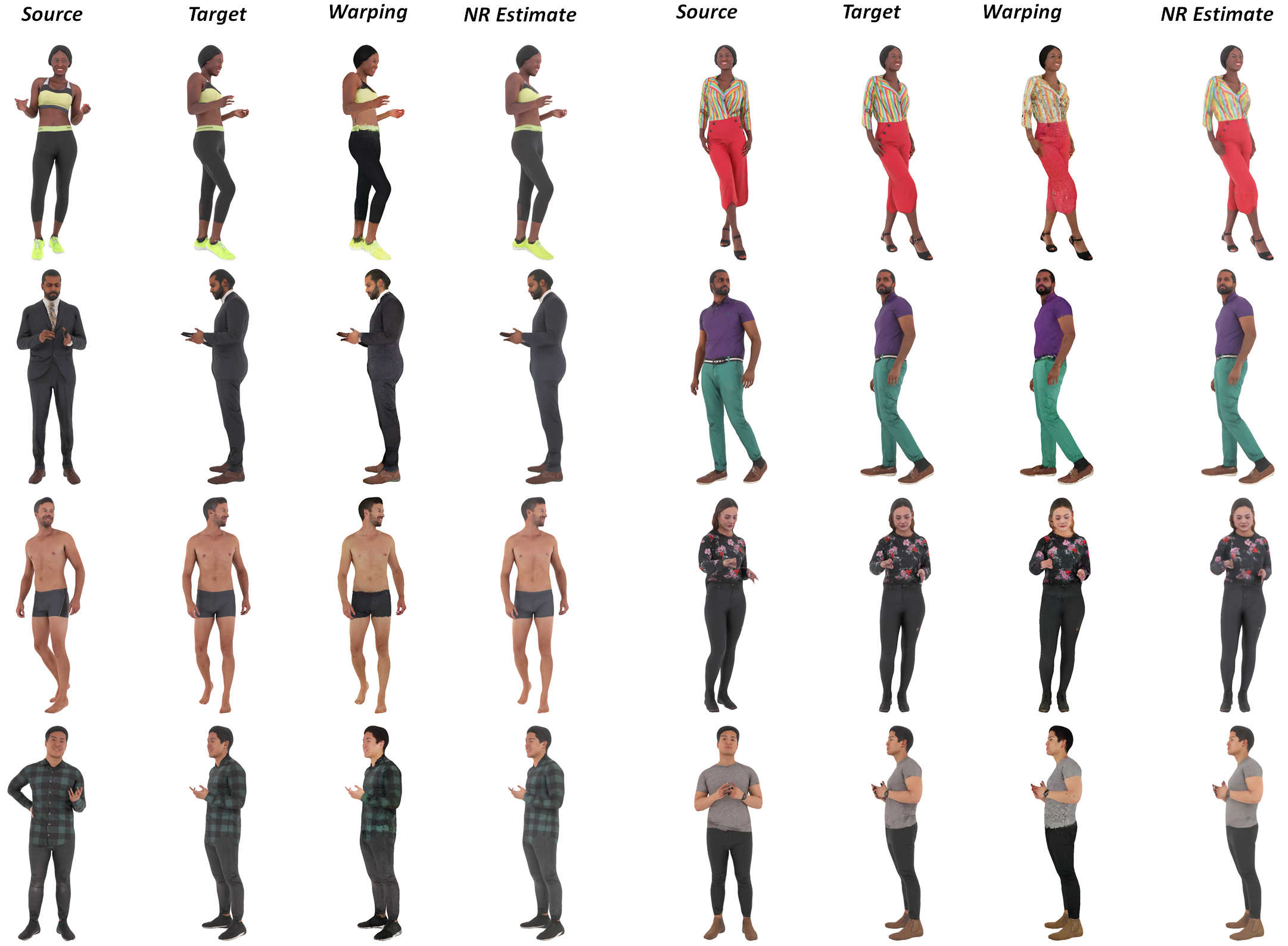}
	\caption{Novel view synthesis results before and after neural re-rendering (NR).}
	\label{fig:sup_rendering}
\end{figure*}

\subsubsection{Neural Re-rendering}
We present an additional step to obtain photo-realistic human renders after texture completion. While we can assume access to 3D geometry for our synthetic data and perform rendering with tools such as Blender Cycles~\cite{blender} or PyTorch3D~\cite{paszke2019pytorch,ravi2020pytorch3d}, this is not the case for real-world examples. One approach to tackle this problem and obtain a 3D geometry would be to estimate the 3D human body pose~\cite{pose3d} and shape from a single image by using any of the recent state-of-the-art methods~\cite{kanazawa2018end, kocabas2020vibe, kolotouros2019convolutional, kolotouros2019learning, pavlakos2018learning, xu2019denserac, zhao2020human, zheng2019deephuman}. However, all these methods estimate the body under the clothing which is usually relatively slimmer and with pose inaccuracies (e.g., the body bends forward) due to the depth ambiguity which makes it unsuitable for rendering the texture of a clothed human on top. Thus, we propose a model for neural re-rendering which aims at learning a function
$\widetilde{X} = R(BodyMap, F_T)$ that given the complete texture map $F_T$ and the estimated BodyMap generates a photorealistic render $\widetilde{X}$ in the image space. The advantage of our approach compared to prior work~\cite{sarkar2020neural} is that BodyMap not only provides a proper silhouette in the image space that needs to be rerendered but also serves as a mapping function between the UV and the image space. This enables us to warp the texture map back to the image space using the BodyMap to obtain an initial estimate which is then fed to an encoder-decoder network that generates the final output render. The fact that BodyMap provides accurate per-pixel foreground estimates makes the warped image well aligned with a target output which simplifies the learning process of the neural renderer. Finally, since during the warping process some texture information can be lost due to warping inaccuracies~\cite{liu2019liquid}, we pass $F_T$ through an encoder network to generate a lower dimensional tensor representation which is then fed via the bottleneck to the decoder of the neural render. We train this network with the same losses described that we used for texture completion network and in addition, we employ the following two losses:

\begin{itemize}
	\item \textbf{Feature Matching Loss.} We use a feature matching loss in the discriminator layers~\cite{isola2017image} to obtain high-frequency details such as wrinkles and cloth patterns which is defined as: $L_{FM} = \sum_{l=1}^3 || D_l(x) - D_l(G(x))||_1$. 
	\item \textbf{Reconstruction Loss in the face region.} Using the segmentation mask of the face region which are then used to employ additional reconstruction losses in that area in order to force the network to estimate more photo-realistic faces and fix artifacts around the eyes and the mouth.
\end{itemize}

\begin{figure*}[t]
	\centering
	\includegraphics[width=\linewidth]{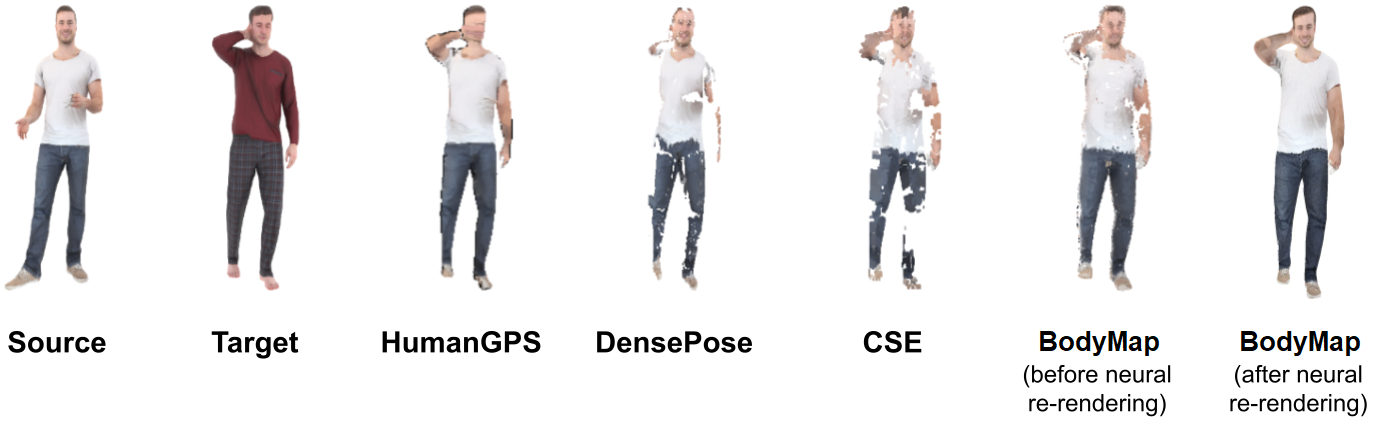}
	\caption{Warping results with BodyMap estimates as a mapping function vs. competitors.}
	\label{fig:warping_new}
\end{figure*}

\subsubsection{Results}
Expanding Figure~4 from the main paper, we present several more results of re-rendered people in~\ref{fig:sup_rendering}. It shows that even before re-rendering (column 6) novel views demonstrate a decent level of details including textile patterns, hairstyles and fingers. Neural re-rendering helps to get rid of occasional artifacts, smooth out the final result and indicate even more fine-grained details.

\subsection{Appearance swapping}

Additionally to generating people in novel views and/or poses, our approach allows to redress people providing renders of them in different clothes. Using BodyMap estimates as a mapping function, we show several examples of appearance swapping in Figure~\ref{fig:appearance_swap}. 

\end{document}